\documentclass[journal,10pt]{IEEEtran}

\usepackage{biblatex}  
\usepackage{graphicx}
\usepackage{caption}
\usepackage{subcaption}
\usepackage{amsmath}
\usepackage{hyperref}

\begin{document}

\captionsetup[figure]{name={Fig.}, labelsep=period, singlelinecheck=off} 

\title{SHLE: Devices Tracking and Depth Filtering for Stereo-based Height Limit Estimation}

\author{

Zhaoxin Fan$^{\ast}$,
Kaixing Yang$^{\ast}$,
Min Zhang,
Zhenbo Song,
Hongyan Liu,
and Jun He

\thanks{$^{\ast}$These authors contribute equally. Jun He is corresponding authors.}
\thanks{Zhaoxin Fan, Kaixing Yang, Min Zhang, and Jun He are with Key Laboratory of Data Engineering and Knowledge Engineering of MOE School of Information, Renmin University of China, 100872, Beijing, China (e-mail: fanzhaoxin@ruc.edu.cn, yangkaixing@ruc.edu.cn, zhanngmin4215@ruc.edu.cn, hejun@ruc.edu.cn).}
\thanks{Zhenbo Song is with School of Computer Science and Engineering, Nanjing University of Science and Technology, 210094, Nanjing, China (e-mail: songzb@njust.edu.cn).}
\thanks{Hongyan Liu is with Department of Management Science and Engineering, Tsinghua University, 100084, Beijing, China (e-mail: hyliu@tsinghua.edu.cn).}
}



\maketitle

\begin{abstract}
Recently, over-height vehicle strike frequently occurs, causing great economic cost and serious safety problems. Hence, an alert system which can accurately discover any possible height limiting devices in advance is necessary to be employed in modern large or medium sized cars, such as touring cars. Detecting and estimating the height limiting devices act as the key point of a successful height limit alert system. Though there are some works research height limit estimation, existing methods are either too computational expensive or not accurate enough. In this paper, we propose a novel stereo-based pipeline named SHLE for height limit estimation. Our SHLE pipeline consists of two stages. In stage 1, a novel devices detection and tracking scheme is introduced, which accurately locate the height limit devices in the left or right image. Then, in stage 2, the depth is temporally measured, extracted and filtered to calculate the height limit device. To benchmark the height limit estimation task, we build a large-scale dataset named ``Disparity Height", where stereo images, pre-computed disparities and ground-truth height limit annotations are provided. We conducted extensive experiments on ``Disparity Height" and the results show that SHLE achieves an average error below than 10cm though the car is 70m away from the devices. Our method also outperforms all compared baselines and achieves state-of-the-art performance. Code is available at \url{https://github.com/Yang-Kaixing/SHLE}.

\end{abstract}

\begin{IEEEkeywords}
Over-height vehicle strike, height estimation, stereo images, devices tracking, depth filtering
\end{IEEEkeywords}

\IEEEpeerreviewmaketitle

\section{Introduction}
\begin{figure}[!t]
\centering
\includegraphics[width=0.49\textwidth]{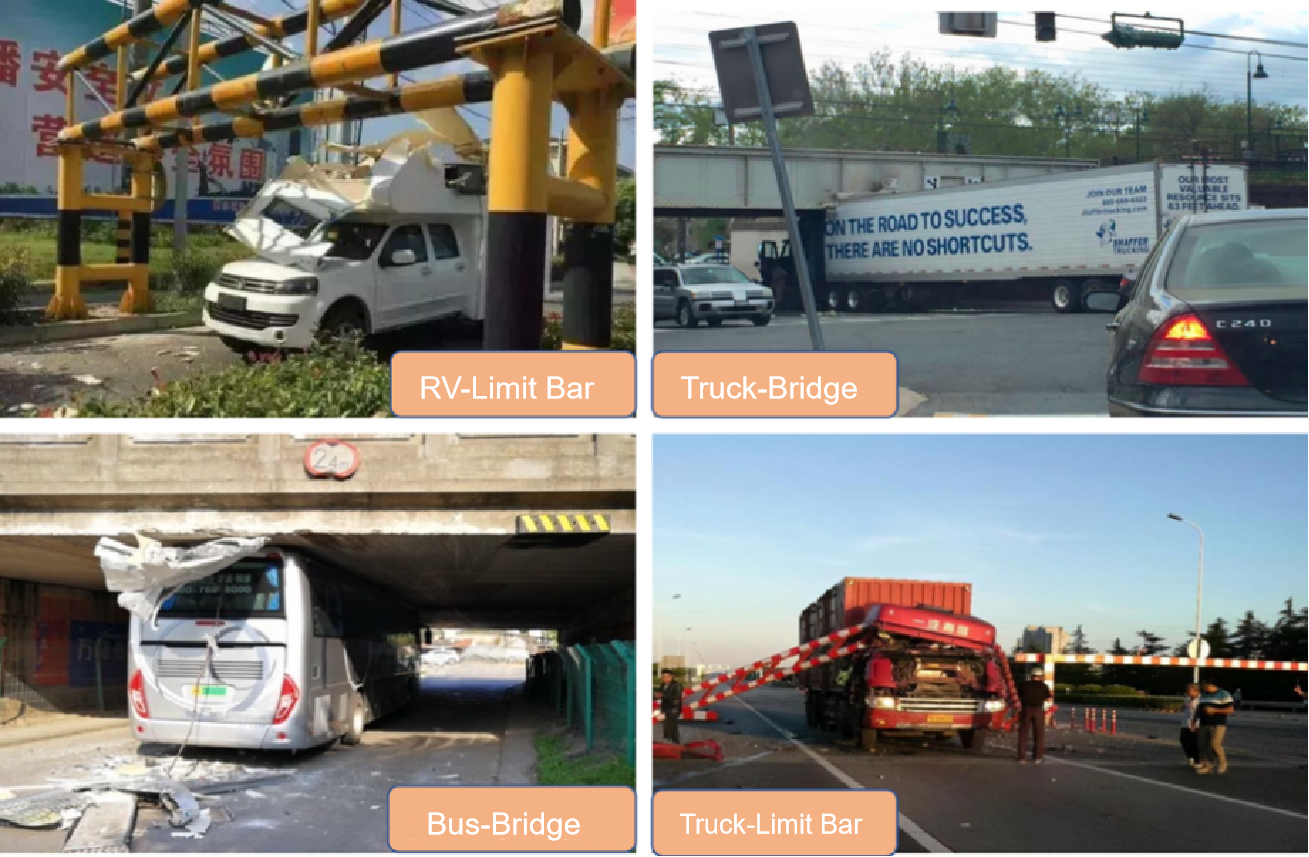}
\caption{Over-height vehicle strike}
\label{Over-height vehicle strike}
\end{figure}
\par
With the development of modernization, different kinds of cars are produced and are running on our roads. Also, with the improvement of people's requirements for travel quality, the shape and size of cars are becoming larger and larger, and the car body is getting higher and higher. While on the other hand, more and more places become to set up some barriers to prevent cars from entering. Height limit devices, for example, is a typical kind of barrier. In our daily life, in addition to the standard height limiting rod, any long strip can be used as a height limit device. For example, a clothes pole or fallen tree. Therefore, height limit devices are frequently viewed in daily life. To this end, the growing number of cars and the ubiquitous height limit devices create a contradiction, i.e., over-height vehicle strike. OHVS is a kind of frequently happen accident as shown in Fig. \ref{Over-height vehicle strike}. The definition of OHVS can be: suppose a car attempts to pass a height limit device while the device is lower than the car's height. In this case, The upper part of the car will collide with this device. We call this kind of collision OHVS \cite{nguyen2018vision}.  To avoid OHVS, an alert system which can accurately discover any possible height limiting devices in advance is necessary to be employed in modern large or medium sized cars. To achieve so, detecting the height limit devices and estimating the heights act as the key of the system. This is a less studied problem because most of existing methods are concentrating on objects in the road, the height limit devices on the sky are often neglected. In this paper, we research the less studied height limit estimation task.

\par
Though being less studied, there still exists some works research how to estimate the height of some objects. Early works explore traditional computer vision technologies to estimate the height limit. For instance,  \cite{bib7} utilizes edge detection and stochastic Hough transform together to detect height limit devices. Though simple, taking single RGB image as input making height estimation an ill-posed problem. To accurately detect and estimate the height of the height limit devices, methods like \cite{bib5, bib6} propose to use LiDAR to capturing point cloud. Though straight forward, LiDAR point cloud is too sparse for accurate height limit estimation. Besides, LiDAR is too expensive for normal users to afford. Recently, some deep learning methods are proposed \cite{simonyan2014very, he2016deep, mobilenetv2} to detect and estimate the height of height limit devices. However, these methods take the Bird Eyes' View as input, which is tailored for aerobat rather than cars.

\begin{figure}
	\subcaptionbox{\centering LIDAR\label{fig: LIDAR}}
	{\includegraphics[width=0.49\linewidth]{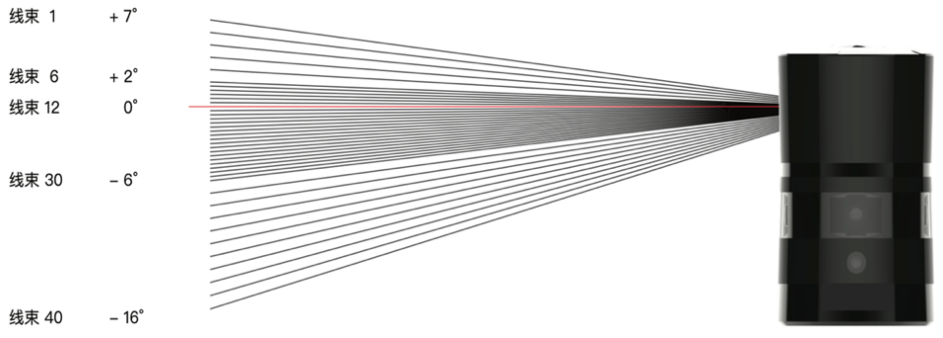}}
	\subcaptionbox{\centering Binocular Camera\label{fig: Binocular Camera}}
	{\includegraphics[width=0.49\linewidth]{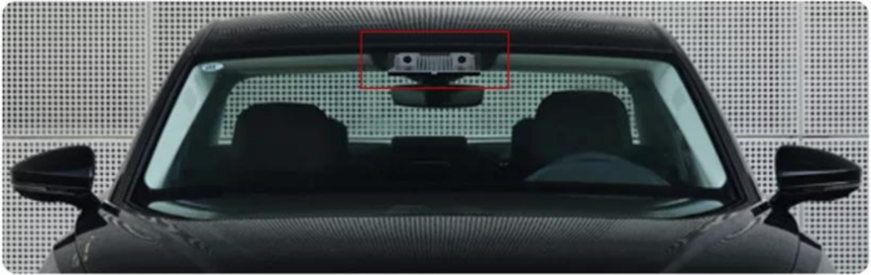}}
	\caption{display effect for cameras}\label{fig: camera}
\end{figure}

To tackle the above issue, we propose a novel stereo-based height limit estimation pipeline named SHLE. In our work, we use stereo cameras to capture left and right images for 3D perception to avoid the ill-posed problem as shown in Fig. \ref{fig: camera}. We choose stereo cameras for the following reasons. 1) stereo cameras is inexpensive compared to devices with LiDAR. So it is possible the deploy it into common cars. 2) Depth reconstructed from stereo images is dense, making accurate height limit estimation being possible. 3)  stereo cameras can perceive scenes far enough away, often more than 200m. This ensures safety. To this end, SHLE makes full use of stereo cameras and make accurate height limit estimation feasible.

\par
In SHLE, the stereo camera is horizontally placed in the front of the car as shown in Fig. \ref{fig: camera}. During driving, the camera captures the left image and right image at each time step. Then, we calculate the sub-pixel level disparity map from the images, which can be then used to recover the depth map. The next pipeline takes the left image and the depth map as input, and output the location and height of the height limit device in the scene. We divide the next pipeline into two stages. In stage 1, a novel devices detection and tracking scheme is introduced, which accurately locates the height limit devices in the left or right image. Then, in stage 2, the depth is temporally measured, extracted and filtered to calculate the height limit bar. SHLE combines deep learning with traditional methods, and each step of our framework is more explainable compared to end-to-end deep learning approaches. In stage 1, a novel devices detection and tracking scheme is introduced, which accurately locates the height limit devices in the left or right image. Then, in stage 2, the depth is temporally measured, extracted and filtered to calculate the height limit device.

\par
Specifically, in stage 1, we firstly execute object detection and object tracking to locate the height limit devices. In particular, to make accurate location, a Faster RCNN \cite{faster_rcnn} detector is employed to detect the height limit devices. Note the detection is not like conventional object detection, which is category-level detection. Our detector detect any category of objects that may be regarded as a height limit devices, e.g., a clothes pole or a fallen tree. This means the detector is category agnostic. After detection, multiple candidate bounding boxes will be generated. To filter redundant targets, we proposed a novel boxes filter rule to select the the one with the highest response as our results. Then, we argue that in a video, not height limit devices in every frame can be well-detected. For those missed framed, we apply object tracking to find possible height limit devices. In this way, once a height limit device appears, we can accurately find it in the image coordinate system.

\par
After that, we come to stage 2. Stage 2 takes the recover depth map and object detection/tracking algorithm as input. A frustum based target extracting scheme is proposed to extract the depth of the target height limit devices, which is back projected into a 3D point cloud. Then, a spatial-based depth filter is utilized to remove the noise points in the candidate point cloud. After that, the height can be easy to be calculated by extract the lower surface of the point cloud. However, we find the height of lower surface is not robust in different frame and the result is highly influenced by ground turbulence and reconstruction error. To tackle the issue, a temporal-based based depth filter is proposed to balance and refine the final estimation results. To this end, we can get accurate height limit estimation.

\par
Note we are the first work of proposing vision based methods for height limit estimation for modern cars. Therefore, there is no public available dataset that we can use. To benchmark our task, we propose a novel large-scale dataset named "Disparity Height".  "Disparity Height" is collected in natural outdoor driving conditions and has a total of 1587 images from 13 scenes, which covers large different types of height limit devices. In this dataset, we not only consider different scenes, but also consider the effects of different lighting, weather, etc. Therefore, "Disparity Height" is a very challenging dataset. It is very meaningful to evaluate the performance of our method and the baselines.

Then, we conduct extensive experiments on our "Disparity Height", and the results show that SHLE achieves an average error below than 10cm though the car is 70m away from the devices. Moreover, we put forward our evaluation metrics, that is, height error and height error rate, for height estimation task. The height error and height error rate of our experiment reached 0.08m and 2.67\% on "Disparity Height" data, which achieves 0.16m and 6.06\% better than the best deep learning based baseline. Hence, our method achieves state-of-the-art performance.

\par
Our contribution can be summarized as:
\begin{itemize}
\item [1)] 
We are the first to propose propose a novel two-stage pipeline named SHLE for stereo-based height limit device detection and height limit estimation, which achieve state-of-the-art performance.
\item [2)] 
In stage 1, we propose a category agnostic object detector and a category agnostic tracker to accurately locate any height limit devices in a video.
\item [3)] 
In stage 2, a frustum based target extractor is proposed to extract 3D height limit devices. Besides, a spatial-based depth filter and a temporal-based height filter are proposed to estimate the height.
\end{itemize}

\section{Related Work}
\subsection{Height Estimation Methods}
\par
Height estimation plays an important role in automated driving field, but there still exists few works research how to estimate the height of some objects. Early works explore traditional computer vision technologies to estimate the height limit. For instance, \cite{bib7} utilizes edge detection and stochastic Hough transform together to detect height limit devices. Though simple, taking single RGB image as input making height estimation an ill-posed problem. To accurately detect and estimate the height of the height limit devices, methods like \cite{bib5, bib6} are proposed to use LiDAR to capture point cloud. Though straight forward, LiDAR point cloud is too sparse for accurate height limit estimation and too expensive for normal users to afford. Besides, LIDAR-based methods do not deal well with complex road conditions, such as flying birds crossing, sharp corners, etc. \cite{ultrasonic1, ultrasonic2} used ultrasonic sensor technology for height estimation of vegetation. The ultrasonic sensors are usually used for data acquisition with the aid of unmanned aircraft systems(UAS), the ultrasonic sensors are not only expensive but also unsuitable for installation on medium or large vehicles. Recently, some deep learning methods are proposed \cite{simonyan2014very, he2016deep, mobilenetv2} to detect and estimate the height of height limit devices. However, these methods take the Bird Eyes' View as input, which is tailored for aerobat rather than cars, and getting the height of a object from a bird's eye view is essentially a depth estimation problem instead of height estimation problem.

\par
In this paper, we design a frustum-based target extractor to obtain 3D information from 2D image inspired by \cite{bib5}. Compared to tradition methods, we design a height limit devices detector and a tracker by deep learning technology, largely reducing the computation and improving the accuracy. Compared to deep learning methods, we are the first to tackle the height estimation task taking front view image as input.  

\subsection{Deep Learning in 2D Vision Field}
\par
Deep learning has achieved excellence in computer vision for 2D data, but the area of height limit estimation is still untouched. Intuitively, we can transfer the methods that have great achievement to our height estimation task. Image classification takes an image as input, and outputs a class. VGG \cite{simonyan2014very}, ResNet \cite{he2016deep}, MobileNet \cite{mobilenetv2, howard2019searching}, etc. are classical image classification networks, and ResNet\cite{he2016deep} has extended to other areas with great performance. Recently, ViT \cite{dosovitskiy2020image} transfer transformer into computer vision area from Natural language process area, and they have got excellent effect \cite{vision_tranformer, swin_tranformer}. Image semantic segmentation takes an image as input, and output a class for each pixel. FCN\cite{long2015fully} is the pioneer of semantic segmentation. U-Net\cite{unet} was proposed right after, and the U-Net family\cite{unet, zhou2018unet++, li2018h} are widely used as the backbone in other task. Monocular depth estimation takes an image as input, and outputs a depth for each pixel. Eigen et al. \cite{eigen2014depth, eigen2015predicting} first proposed a multi-scale network framework based on encoder-decoder. DORN\cite{dorn} proposed a novel backbone that achieved first place on several datasets. However, recently more works focus on binocular vision, few significant breakthroughs\cite{bhat2021adabins} have been made recently in monocular area. Object Detect takes an image as input, and outputs a candidate bounding box set. We usually divide object detection into two types: 1) one-stage. 2) two stage. The two-stage method is highly accurate but slow, it uses a two-stage cascade to fit the bounding box, with coarse regression followed by fine tuning \cite{faster_rcnn, fcos}. The one-stage method is fast but slightly less accurate, it skips suggested bounding boxes stage and directly produces the class and bounding box of the object \cite{retinanet, yolov5, centernet}. Object tracking takes a video as input, and finds out which parts of the image depict the same object in different frames\cite{boosting, mil, csrt, kcf, moose, medianflow}, and we often use detectors as starting points.

\par
In our paper, SHLE considers the height estimation task as a regression problem at image level. In our work, we adopt object detection and tracking for height limit devices location. However, we conduct category agnostic estimation for objects in the sky, while the above mentioned methods all focus on category-level detection for object on road.  Moreover, the above mentioned models are all end-to-end methods, while SHLE makes our task two stage: stage 1 makes accurate location, and stage 2 obtains refined heights, which make our SHLE more explainable and accurate. 

\subsection{Deep Learning in 3D Vision Field}
\par
Recent years have witnessed rapid progress in deep learning for 3D vision, spanning both 3D perception and 3D content generation. 

\par
On the perception side, 3D understanding has evolved from basic point-set learning to increasingly sophisticated geometric reasoning and scene-level perception. Early point-based methods such as PointNet~\cite{qi2017pointnet} first demonstrated that neural networks can directly process unordered point sets through shared point-wise feature extraction and permutation-invariant aggregation. Building on this idea, PointNet++~\cite{qi2017pointnet2} introduced hierarchical neighborhood grouping and multi-scale local feature learning, significantly improving the modeling of fine-grained geometric structures. Subsequent works further strengthened local context modeling from different perspectives. For example, DGCNN~\cite{wang2019dynamic} dynamically constructed local graphs in feature space to capture edge-based geometric relations, while PointCNN~\cite{li2018pointcnn} explored convolution-like operators on irregular point sets to better exploit local spatial correlations. In parallel, voxel-based and hybrid representations became important for large-scale 3D scene understanding, especially in autonomous driving and detection tasks. VoxelNet~\cite{zhou2018voxelnet} pioneered end-to-end 3D object detection from raw point clouds by learning voxel-wise representations directly from LiDAR data. PointPillars~\cite{lang2019pointpillars} further improved efficiency by organizing points into vertical columns and converting point cloud encoding into a fast pseudo-image representation. Frustum PointNets~\cite{qi2018frustum} combined 2D detection and 3D point cloud reasoning through frustum-based extraction, showing the effectiveness of integrating image cues with point-based 3D understanding. PointRCNN~\cite{shi2019pointrcnn} then demonstrated that high-quality 3D proposals can be generated directly from raw point clouds in a bottom-up manner, while VoteNet~\cite{qi2019votenet} introduced deep Hough voting to support robust object detection in 3D indoor scenes. More recently, PV-RCNN~\cite{shi2020pvrcnn} combined voxel CNNs and point-based set abstraction to better balance efficiency and accuracy, and transformer-based architectures such as Point Transformer~\cite{zhao2021pointtransformer} further improved long-range contextual modeling and dense prediction on point clouds.

\par
Beyond 3D perception, 3D generative modeling has also advanced rapidly. For 3D objects, DreamFusion~\cite{poole2022dreamfusion} demonstrated that pretrained 2D diffusion models can serve as powerful priors for text-to-3D synthesis, opening a new direction for leveraging 2D generative models in 3D creation. GET3D~\cite{gao2022get3d} further showed that high-quality textured 3D meshes can be generated from 2D image collections, while Shap-E~\cite{jun2023shapee} improved efficiency by learning conditional implicit 3D functions. More recently, large reconstruction/generation models such as LRM~\cite{hong2024lrm} have pushed single-image-to-3D generation toward a more scalable and generalizable paradigm. A related yet distinct branch is human motion generation, which can be categorized according to different conditioning signals~\cite{zhang2025robust}. First, \emph{text-to-motion} methods aim to synthesize motion sequences from natural language descriptions, with representative works including T2M-GPT~\cite{zhang2023generating}, MDM~\cite{tevet2023human} and DynMask~\cite{zhou2026not}. Second, \emph{speech-to-face} methods focus on generating speech-driven facial motion or talking faces from audio, as explored in Speech2Face~\cite{oh2019speech2face}, FaceFormer~\cite{fan2022faceformer}, EmoTalk~\cite{peng2023emotalk}, SelfTalk~\cite{peng2023selftalk}. Third, \emph{music-to-dance} methods generate dance motions aligned with rhythm and musical structure~\cite{yang2024beatdance,yang2025matchdance,yang2026tokendance,yang2025mace}, where Bailando~\cite{siyao2022bailando}, EDGE~\cite{tseng2023edge}, MEGADance~\cite{yang2025megadance}, and FlowerDance~\cite{yang2025flowerdance} are representative examples. Fourth, \emph{speech-to-gesture} methods synthesize co-speech body gestures from audio or speech signals, with recent progress represented by Semtalk~\cite{zhang2025semtalk}, Echomask~\cite{zhang2025echomask} and HoloGest~\cite{zhang2026mitigating}. Finally, \emph{group motion generation} considers the synthesis of coordinated motions involving multiple individuals, where the model must capture both individual motion dynamics and inter-person interactions, including DualTalk~\cite{peng2025dualtalk}, CoDancers~\cite{yang2024codancers}, CoheDancers~\cite{yang2024cohedancers}.

\par
In this paper, unlike previous methods that take depth map or point cloud as an input of a deep network, instead, after locating the height limit device area, we direly calculate the height of the selected points from the extracted point cloud. However, some errors avoidably occur may occur, purely choose the lowest height of the selected may be deeply influenced. Thus, we additionally create a spatial-based depth filter and temporal-based height filter to refine our result.

\section{Problem Definition}
\par
We design a pipeline named SHLE that combines deep learning and traditional methods. For each frame, SHLE directly consumes a RGB image $I \in \mathrm{R}^{W \times H \times 3}$ and its disparity map $D_{i} \in \mathrm{R}^{W \times H \times 1}$ as inputs, where $W$ and $H$ are the width and height of $I$ or $D_{i}$. The outputs of SHLE are the height limit device's height $h_{i} \in h_{tf}$ for each image $I$ and height $h$ for a scene. We note each scene contains $N$ frames in a video. SHLE is a two-stage pipeline, we will define it in turn.

\par
In stage 1, we define height limit device detector as $F(*)$ that can generate a bounding box to locate height limit device for each frame $I$, i.e., $ I \in \mathrm{R}^{W \times H \times 3} \rightarrow b \in \mathrm{R}^{4 \times 1}$, where $b$ includes width, height, horizontal and vertical coordinates of the rectangular vertex. Then we define height limit device tracker as $G(*)$ that can complete the missing bounding box in some frames, i.e., $ b_{i} \in \mathrm{R}^{m \times 4 \times 1} \rightarrow b_{j} \in \mathrm{R}^{N \times 4 \times 1}$, where $b_{i}$ is the missing bounding box set and $m$ is the number of detected frames by height limit device detector,  $b_{j}$ is the completed bounding box set and $N$ is number of all the frames in the scene.

\par
In stage 2, we first define the disparity-to-depth transfer function as $B(*)$, i.e., $D_{i} \rightarrow D_{e} \in R^{W \times H \times 1}$, where $D_{i}$ is the disparity map and $D_{e}$ is the depth map. Secondly, for each frame, we define the frustum based target extractor as $H(*)$ that can project the points in bounding box $b$ from pixel coordinate system to world coordinate system, i.e., $p_{p} \in \mathrm{R}^{n \times 2} \rightarrow p_{w} \in \mathrm{R}^{n \times 3}$, where $n$ is the number of the points in bounding box, $p_{p}$ is the points set in 2D $(x, y)$ and $p_{w}$ is the one in 3D$(x, y, z)$. Thirdly, we define a spatial-based depth filter as $P(*)$ that can filter the noise points in 3D point cloud, i.e., $p_{w} \in \mathrm{R}^{n \times 3} \rightarrow p_{df} \in \mathrm{R}^{k \times 3}$, where $n$ is the number of the points unfiltered and $k$ is the number of remained points after filtering. Finally, we define a temporal-based height filter as $Q(*)$ that make our predictions in a same scene more smoother and stable, i.e., $h_{df} \in \mathrm{R}^{N \times 3} \rightarrow h_{tf} \in \mathrm{R}^{N \times 3}$, where $N$ is the number of all frames in a scene, $h_{df}$ is the height prediction set of all frames and $h_{tf}$ is the refined height prediction set. The height $h$ for a scene can be calculated by averaging all the heights in $h_{tf}$.

\section{Method}
\begin{figure*}[!t]
  \centering
  \includegraphics[width=0.98\linewidth]{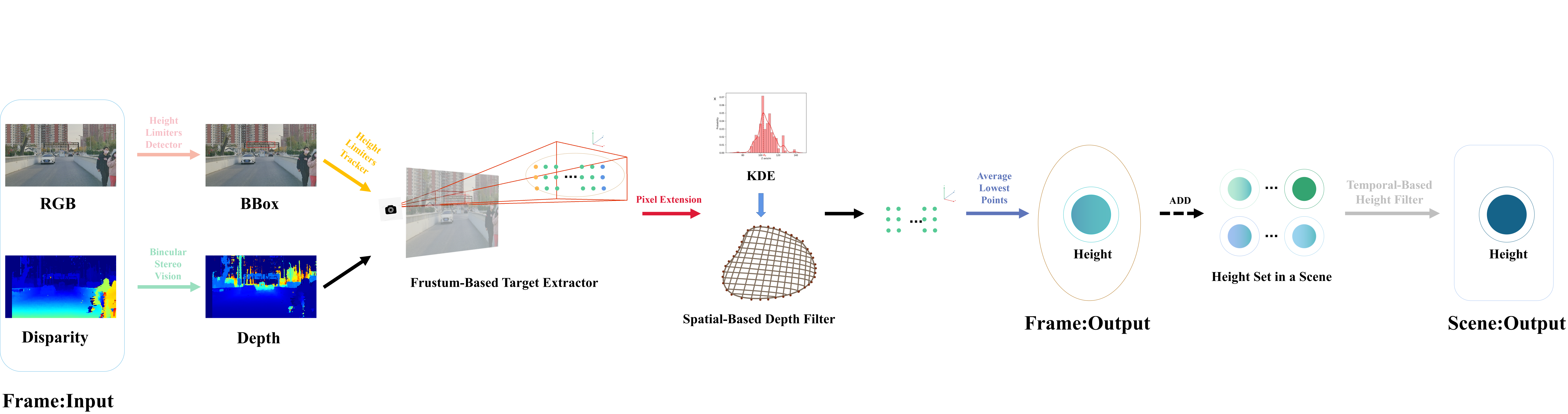}
  \caption{SHLE. For each frame $f_{i}$, SHLE takes disparity map $D_{i}$ and RGB image $I$ as input and outputs the height $h$ of corresponding height limit device. For each scene, SHLE will generate a scene-level height after collecting all frames' output. In stage1, for each frame, we firstly execute object detection by Height Limit Device Detector $F(*)$ to a get bounding box $b$ of RGB image $I$, secondly apply object tracking by Height Limit Device Tracker $G(*)$ to complete missing frames in a scene, and execute Pixel Extension for $b$. In stage2, we firstly transfer the disparity map into depth map through Binocular Stereo Vision $B(*)$, secondly obtain the 3D point cloud by frustum-based target extractor $H(*)$, thirdly refine the points by spatial-based depth filter $P(*)$, finally execute Averaging Lowest Points to get the frame's height. After collection all frames' output, we utilize a temporal-based height filter $Q(*)$ to get the adjusted height $h_{i} \in h_{tf}$ for each frame and the averaged height $h$ for the scene.}
  \label{fig: framework}
\end{figure*}

\par
As shown in Fig. \ref{fig: framework}, our SHLE consists of two stage: 1) Height Limit Devices Detector and Tracker. 2) Frustum-Based Target Extractor, Spatial-Based Depth Filter and Temporal-Based Height Filter. We will introduce each module in turn. SHLE is more accurate and explainable compared to the end-to-end deep learning methods, and SHLE is less computationally costly and cheaper in application. Specifically, our SHLE firstly apply a category agnostic detector and a category agnostic tracker to accurately locate any height limit devices in a video. Secondly, we utilize a frustum-based target extractor which can extract the points belonging to 3D height limit devices from 2D pixel coordinate system to 3D world coordinate system. Thirdly, to remove the noise, we design a spatial-based depth filter that effectively mines the significant difference between height limit devices and nearby noise. At frame level, we can obtain the height by average the lowest points after depth filtering. At scene level, after collecting all the frames' outputs in the same scene, we create a temporal-based height filter that can weaken the effects of road bumps and experimental errors, then the height of the scene can be generated by averaging all the frames' output after height filtering. We will introduce above step in turn.

\subsection{Stage 1: Height Limit Devices Detector and Tracker}

\par
Stage 1 takes the RGB image as input, and outputs the bounding box of height limit devices in RGB image. If one chooses a neural network based end-to-end approach to solve the height estimation problem, the intuitive idea is to design an encoder to extract features from the image and then perform linear regression to output the predicted height. For height estimation problem, it will bring much unnecessary noise, the height limit device part we need to focus on only merely accounts for a small portion of the overall image. However, above end-to-end approach focus more on the overall picture, and irrelevant parts of the rest image sometimes negatively affect the results, and the same scene prediction will be unstable on the validation set. To solve these problems, we introduce a height limit devices detector and tracker in stage 1. 

\subsubsection{Height Limit Devices Detector}
\par
Our height limit devices detector takes a RGB image as input, and outputs the location and confidence of candidate bounding boxes. Unlike the semantic segmentation task, the annotation of object detection task is simpler, which make experimental cost lower. Moreover, when detection for height limit devices, there is a significant difference between the effective and noisy point parts within the bounding box. A spatial-based depth filter is designed later on which can effectively eliminate the noise and greatly improve its accuracy, comparable to semantic segmentation method.

\par
2D object detection techniques have performed outstandingly in various scenes, many excellent models have emerged across the board. In our task, the input video is taken from far to near in different outdoor driving scenes, and height limit devices only take up a small part of images. Thus, detection for height limit device is belong to small object detection and multi-scale detection task.

\par
Meanwhile, Faster RCNN\cite{faster_rcnn} has achieved significant experimental results for solving multi-scale and small objection detection problems, and got outstanding performance on multiple datasets. Faster RCNN\cite{faster_rcnn} is faster than the other two-stage models and has higher accuracy than the other one-stage models. Thus, We decided to apply Faster RCNN\cite{faster_rcnn} to accomplish our task after research and analysis, brief description of its architecture is as follows.

\par
Faster RCNN\cite{faster_rcnn} can actually be divided into 4 main part: Conv Layers(CL), Region Proposal Networks(RPN), RoI Pooling, Classification and BBox Prediction. CL is used to extract feature maps, and extracted feature maps are shared for subsequent RPN and fully connected layers. RPN is used to generate regional proposals. RoI pooling collects above feature maps and proposals, and outputs proposal feature maps. with multiple kernels, RoI pooling makes proposal feature maps a fixed size even if from variable one, which makes Faster RCNN\cite{faster_rcnn} adapted for multi-scale detection problem. Classification and BBox Prediction uses proposal feature maps to calculate the class of region proposal, and again executes bounding box regression to obtain the exact final position of detection box. Note, Faster RCNN\cite{faster_rcnn} is originally designed for category-level object detection, while we extend it to category agnostic height limit device detection.

we not simply use faster-RCNN for detection. 

\subsubsection{Novel Boxes Filter Rule}

\begin{figure}[!h]
    \centering
    \subcaptionbox{\centering confidence highest\label{fig: confidence}}
    {\includegraphics[width=0.48\linewidth]{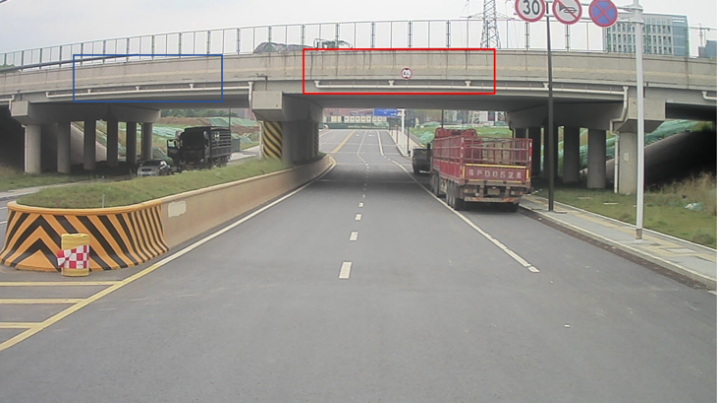}}
    \subcaptionbox{\centering too high\label{fig: high}}
    {\includegraphics[width=0.48\linewidth]{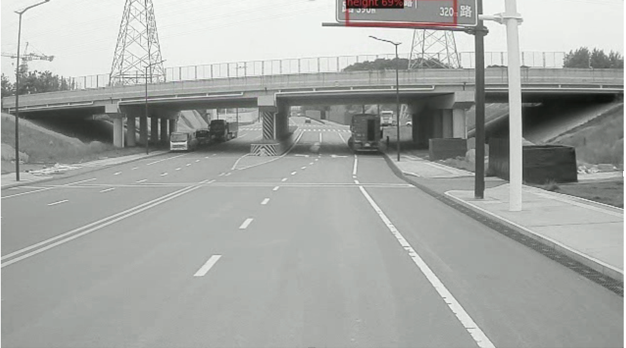}}
    \subcaptionbox{\centering too close\label{fig: close}}
    {\includegraphics[width=0.48\linewidth]{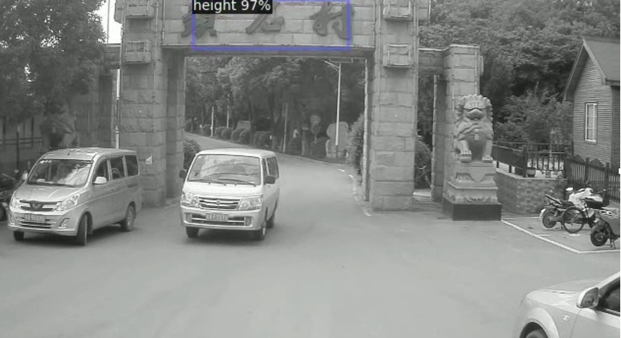}}
    \subcaptionbox{\centering too low\label{fig: too low}}
    {\includegraphics[width=0.48\linewidth]{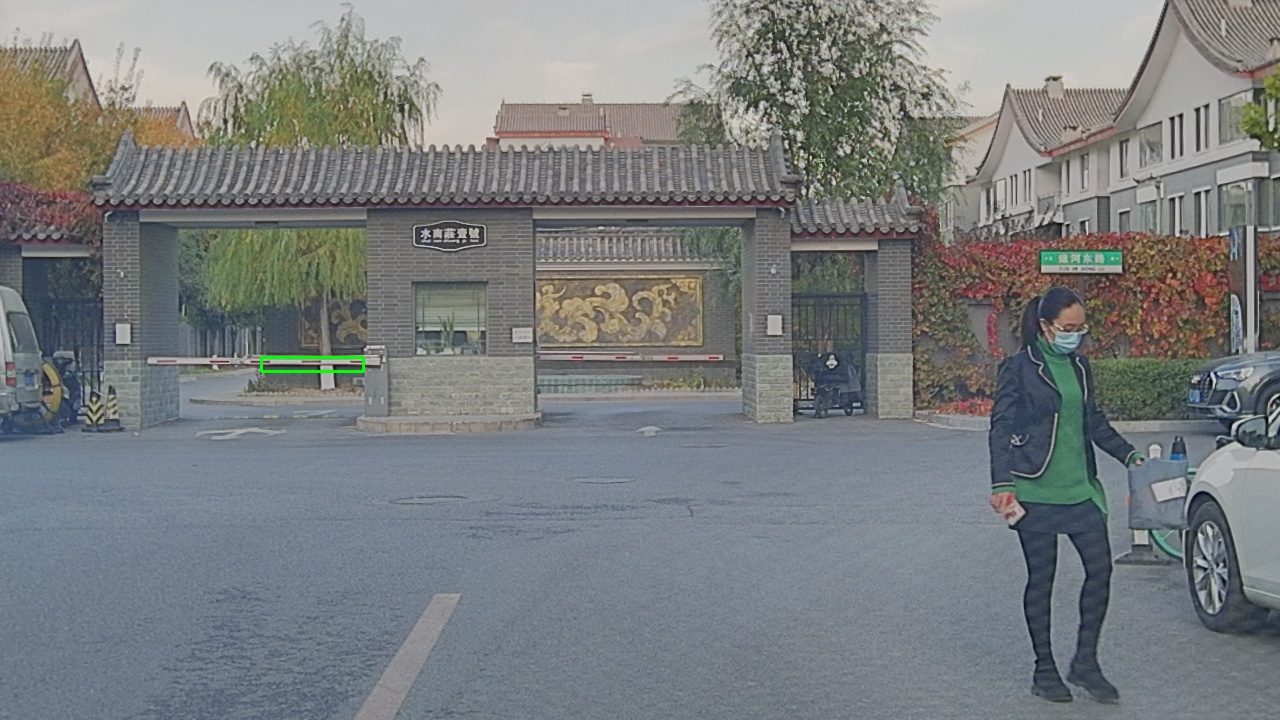}}
    \caption{Compare between Predict and Ground Truth}
        \label{filter}
\end{figure}

\par
After object detection, multiple candidate bounding boxes are generated, but we only need one for our task. General process is to treat the one with highest confidence as the final output. However, the one highest confidence score does not always mean the best one in our task, as show in Fig. \ref{fig: confidence}, where the left blue box has the height confidence score but right red box is ground truth. Therefore, we propose a novel detection box filter rule, exclusively for height estimation task in outdoor driving scenes, shown as follows.

\par
Firstly, we select detected boxes that locates relatively right. According to the local road traffic rules where we collect the data, vehicles are driving on the right, we use a left-hand drive vehicle for data acquisition, and height limit device directly in front of the vehicle is located in the right half of the acquired image. Thus, the rights in candidate boxes set are seemingly more reliable. In this way, we remove the 50\% leftest ones from the set. 

\par
Secondly, we select detection boxes that locates relatively high. Due to the diversity of height limit device, many other objects share a high degree of similarity with height limit device, such as steps, railings, etc., as shown in Fig. \ref{fig: too low}. Therefore, false positives sometimes may occur. Moreover, height limit device should be at least 2m tall, hence they commonly disappear on the relatively top of the image. To this end, we should only keep candidate bounding boxes that lay on the top of the image. Thus, we remove the 50\% lowest ones from the set. 

\par
Thirdly, we remove detected boxes that exceeds the top edge of the image. There are two cases that correspond to it: 1) Some height limit devices that do not affect the normal movement of vehicles are also detected during object detection, such as billboards, power lines and traffic lights, and they tend to appear in the top edge of the image, as in Fig. \ref{fig: high}. 2) When the vehicle is particularly close to the effective height limit device, it will also appear in the top edge of the image, but it is obviously no longer necessary to make height estimations, as in Fig. \ref{fig: close}. Thus, we remove the boxes that exceeds the top edge of the image from the set of alternative prediction boxes.

\par
Lastly, we choose the detected box that gets the highest confidence score among the remaining candidate boxes set.

\subsubsection{Height Limit Devices Tracker}

\par
we argue that in a video, not height limit devices in every frame can be well-detected by the detector. For those missed frames, we apply object tracking to find possible height limit devices.

\begin{figure}[!h]
  \centering
  \includegraphics[width=0.49\textwidth]{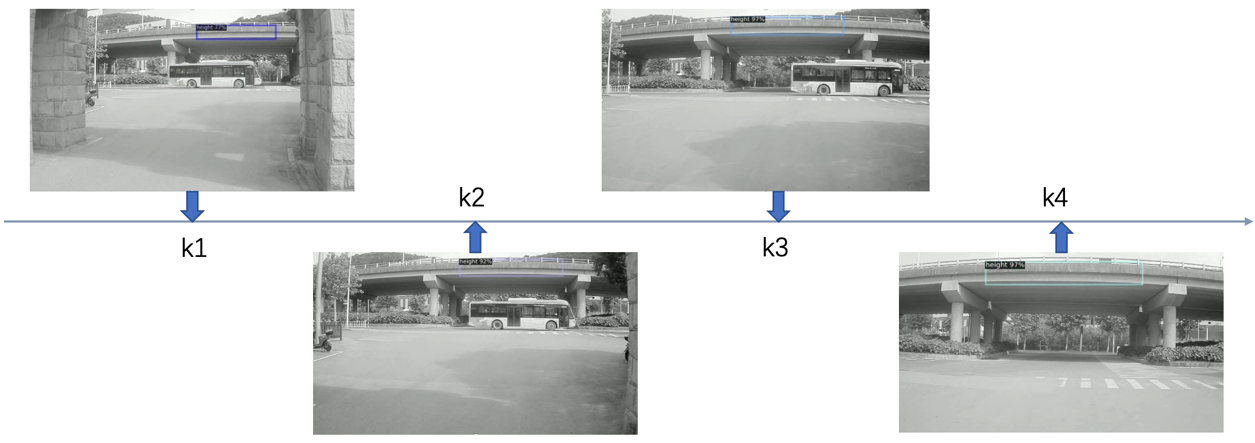}
  \caption{Object Tracking}
  \label{Object Tracking}
\end{figure}

\par
For example, our data set is a collection of image sequences taken in different scenes. Suppose a scene contains a total of M frames of image sequences with valid height limit devices, but the images with valid prediction boxes detected by object detection method may be less than M frames, as shown in Fig. \ref{Object Tracking}, the object detection frames at $k1, k2, k3, k4$ have output, but in the period of $k1 \rightarrow k2, k2 \rightarrow k3, k3 \rightarrow k4$ have no output. Therefore, we need to use object tracking, such as KCF\cite{kcf}, CSRT\cite{csrt}, MOSSE\cite{moose}, MIL\cite{mil}, etc., to fill in the blank frames. For example, we treat k1 as the first frame for object tracking of $k1 \rightarrow k2$, $k2 \rightarrow k3, k3 \rightarrow k4$ are as well.

\par
In our work, we adopt multiple instance learning (MIL) tracker \cite{mil} as height limit device tracker. The object tracking type in this paper is belonging to online tracking, and the main problem in online tracking is "drift". Generally, low accuracy of the samples used in the classifier update is the main reason of "drift". Conventional object tracking methods have three components: appearance model, motion model, and a search strategy for finding. To solve this problem, MIL\cite{mil} focuses on the first component. Since the accuracy of samples is questionable, MIL \cite{mil} expands the "sample" set into a "bag" set, and then updates the object location and classifier by selecting the time with the lowest error rate. In this way, it improves the accuracy and speed with few parameters.

\subsection{Stage2: Height Limit Device Extraction and Filtering}

\par
Stage 2 takes the recover depth map and the bounding boxes generated by object detection/tracking algorithm as input, and then output the height of corresponding height limit device. In stage 2, the depth is temporally measured, extracted and filtered to calculate the height limit device, we will present it in turn.

\begin{figure*}
	\centering
	\subcaptionbox{\centering Binocular Stereo Vision\label{fig: Binocular Stereo Vision}}
	{\includegraphics[width=0.32\linewidth]{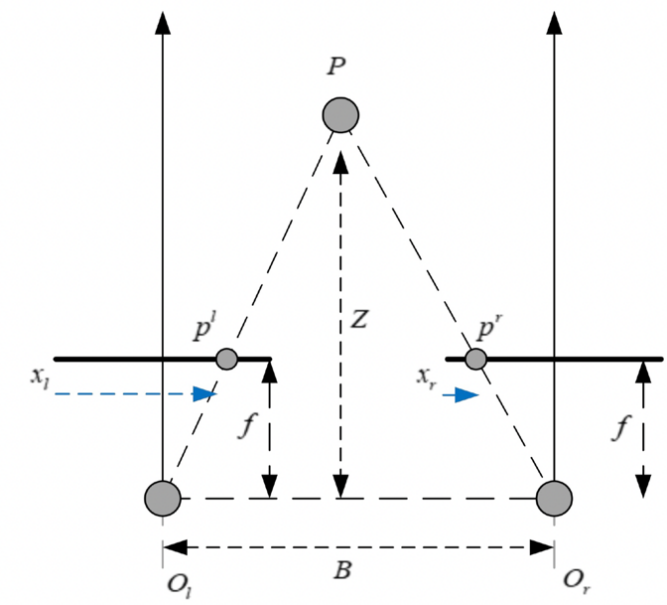}}
	\subcaptionbox{\centering World to Pixel\label{fig: World to Pixel}}
	{\includegraphics[width=0.32\linewidth]{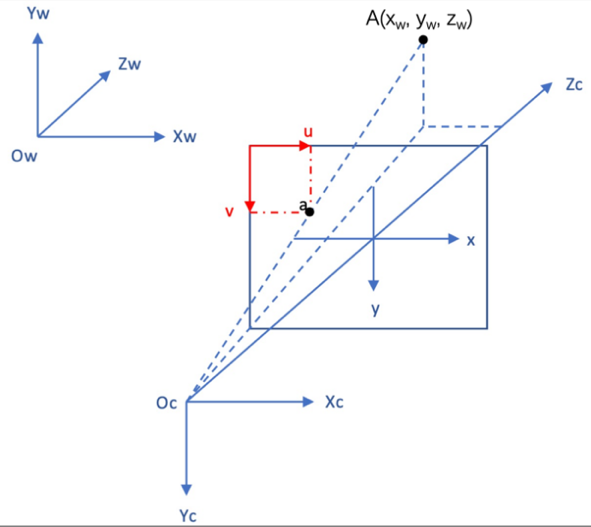}}
	\subcaptionbox{\centering Reference Image\label{fig: Reference Image}}
	{\includegraphics[width=0.32\linewidth]{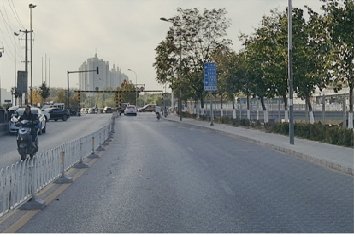}}
	\subcaptionbox{\centering Target Image\label{fig: Target Image}}
	{\includegraphics[width=0.32\linewidth]{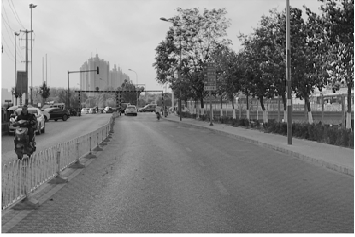}}
	\subcaptionbox{\centering Disparity Map\label{fig: Disparity Map}}
	{\includegraphics[width=0.32\linewidth]{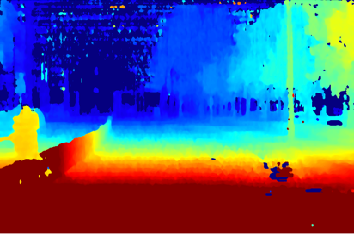}}
	\subcaptionbox{\centering Depth Map\label{fig: Depth Map}}
	{\includegraphics[width=0.32\linewidth]{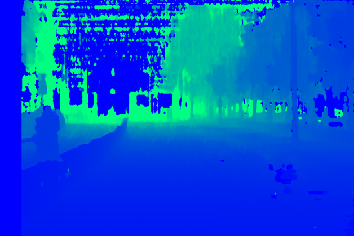}}
	\caption{Frustum}\label{Frustum}
\end{figure*}

\subsubsection{Depth Acquisition}
\par
Stereo vision systems usually employ two identical cameras to simulate the human eyes, one of which is the target camera as Fig. \ref{fig: Target Image} and the other is the reference camera as Fig. \ref{fig: Reference Image}. The simultaneous acquisition of binocular stereo cameras enables the acquisition of stereo image pairs of the same scene, followed by stereo calibration, stereo matching, and finally we acquire depth from disparity by binocular stereo camera geometric model.

\par
Specifically, we can get depth information from disparity information by camera geometric model as Fig. \ref{fig: Binocular Stereo Vision} shows. $O$ and $O^{\prime}$ are the optical centers of the left and right cameras, and line $B$ is the baseline of the binocular camera used. In this paper, $B$ is set to 120 mm. The focal length of the two cameras represented as $f$. The point $P$ in 3D coordinates on the left and right camera imaging planes are respectively $p^{l}$ and $p^{r}$, they are guaranteed on the same line of pixels. $x^{l}$ and $x^{r}$ denote the distance from left boundary of the imaging plane to point respectively on the left and right camera image planes. The depth of $P$ is $Z$, which can be calculated by properties of similar triangles as Eq. \ref{disparity2depth1}.
\begin{equation}
\begin{array}{l}
\displaystyle{Z=\frac{B f}{\left|x^{l}-x^{r}\right|}}
\label{disparity2depth1}
\end{array}
\end{equation}

\subsubsection{Frustum-Based Target Extractor}

\par
After obtaining the depth map $d_{e}$ from disparity map $d_{i}$ by Eq. \ref{disparity2depth1} and the bounding box $b$, we design a frustum-based target extractor to extract the points belonging to target height limit devices and then back projected into a 3D point cloud. 

\par
Firstly, it is necessary to understand the process of mapping world coordinates system to an pixel coordinates system. Suppose a point $A\left(x_{w}, y_{w}, z_{w}\right) \in p_{w}$ in the world coordinate system is mapped to a point $a(u, v) \in p_{p}$ on an image, as shown in Fig. \ref{fig: World to Pixel}. Formal representation is shown in Eq. \ref{world2pixel}, $f_{x},f_{y},c_{x},c_{y}$ are the known camera internal parameters, $R$ and $T$ can be obtained from camera external parameters.

\begin{equation}
\displaystyle
z_{c}\left(\begin{array}{c}
u \\
v \\
1
\end{array}\right)=\left[\begin{array}{ccc}
f x & 0 & c_{x} \\
0 & f y & c_{y} \\
0 & 0 & 1
\end{array}\right]\left[\begin{array}{ll}
R & T
\end{array}\right]\left[\begin{array}{c}
x_{w} \\
y_{w} \\
z_{w} \\
1
\end{array}\right]
\label{world2pixel}
\end{equation}

\par
Secondly, we select the points $p_{p}$ in bounding box $b$ and project points from 2D to 3D. Based on the Eq. \ref{world2pixel}, we can derive the transformation equation Eq. \ref{pixel2camera} from a point $a(u, v) \in p_{p}$ in pixel coordinate system to corresponding point $A^{\prime}\left(x_{w}, y_{w}, z_{w}\right) \in p_{c}$ in camera coordinate system. As Fig. \ref{fig: World to Pixel} shown, the projection process from 2D points to 3D points can be treated as a frustum, i.e., the recovered 3D points can be included by a frustum, where the optical center of the camera is frustum's origin, the bounding box in pixel plane is frustum's near clipping plane, and the plane which contains the point with highest depth and is parallel pixel plane is frustum's far clipping plane.

\begin{equation}
\displaystyle
\left\{\begin{array}{l}
x_{c}=z_{c} \cdot\left(u-c_{x}\right) / f \\
y_{c}=z_{c} \cdot\left(v-c_{y}\right) / f \\
z_{c}=z_{c}
\end{array}\right.
\label{pixel2camera}
\end{equation}

\par
Thirdly, we project points in camera coordinate system to world coordinate system. Because a real world static point in camera coordinate system with different position or pose will lead different coordinate, we need to project the point to the world coordinate system, which is independent of camera pose and position. Based on the Eq. \ref{world2pixel}, we can derive the transformation equation Eq. \ref{camera2world} from point $A^{\prime}\left(x_{w}, y_{w}, z_{w}\right)$ in the camera coordinate system to point $A\left(x_{w}, y_{w}, z_{w}\right)$ in world coordinate system.

\begin{equation}
\displaystyle
\left[\begin{array}{l}
\mathrm{x}_{w} \\
\mathrm{y}_{w} \\
\mathrm{z}_{w}
\end{array}\right]=\mathrm{R}^{-1}\left(\left[\begin{array}{l}
\mathrm{x}_{c} \\
\mathrm{y}_{c} \\
\mathrm{z}_{c}
\end{array}\right]-\mathrm{T}\right)
\label{camera2world}
\end{equation}

\par
Finally, we add the mounting height to points' y-axis. However, $y_{w}$ at point $A \in p_{w}$ does not yet represent the real world height. $y_{w}$ only represents the relative height of $A$ and stereo camera. At this point, it is also necessary to know the mounting height of stereo camera $H_{m}$, and $H_{m}$ is 1.45m in this paper. So y-axis value $y^{\prime}_{w}$ of the point $A$ in 3D space should be: $y^{\prime}_{w} = y_{w} + H_{m}$.

\par
Since we have already obtained the real world coordinate information of the points belonging to height limit device by our frustum-based target extractor. Intuitively, we can get the corresponding height by directly calculating the smallest value in the y-axis of the point cloud. However, there are some unavoidable errors and noise occurring in some previous steps, such as object detection, depth acquisition, etc.. Therefore, we create a spatial-based depth filter and a temporal-based height filter to make refinement.

\subsubsection{Spatial-Based Depth Filter}
\par
\begin{figure}[!h]
	\centering
	\subcaptionbox{\centering \label{fig: Incorrect Contain A}}
	{\includegraphics[width=0.49\linewidth]{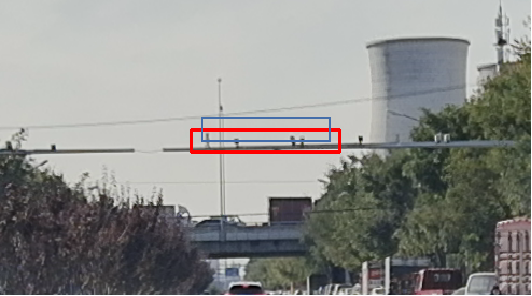}}
	\subcaptionbox{\centering \label{fig: Incorrect Contain B}}
	{\includegraphics[width=0.49\linewidth]{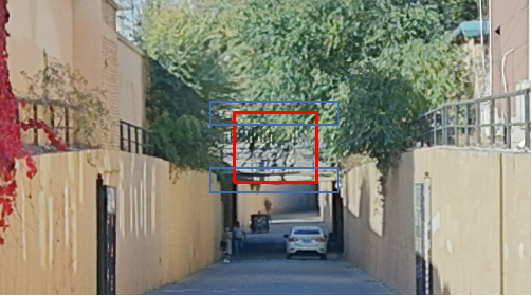}}
	\caption{Incorrect Contain. Red bounding box is our labeled ground truth, and the blue bounding box is incorrect contain part.}\label{fig: incorrect contain}
\end{figure}

To remove the noise points in the candidate point cloud, we design a spatial-based depth filter. For example, although stereo camera can provide us a dense map, however, it is sensible to lightness and prone to mismatch in the low-texture part, few depth errors will inevitably occur sometimes. At the same time, since object detection differs from semantic segmentation, object detection bounding boxes we predict will often inevitably contain a small portion of irrelevant noise, such as sky, floor, trees etc., as shown in Fig. \ref{fig: incorrect contain}. All above will deeply influence the output height.

\par
Although the noise are extremely close to the height limit device in pixel coordinate system, they still have a large gap in depth, for instance, the gap between height limit device and sky. Moreover, in outdoor driving scenes for image acquisition, majority of the time the binocular cameras on vehicle are directly in front of the height limit device, surely, there will still exist few slight angle deviations. Therefore, we can consider that the effective depth of each point of the same height limit device point cloud is highly approximate, at least, keeps stable in a small interval. Thus, a we can easily divide the negative and positive part through a spatial-based depth filter.

\begin{figure}[!h]
	\centering
	\subcaptionbox{\centering Before Extension\label{fig: Before Extension}}
	{\includegraphics[width=0.49\linewidth]{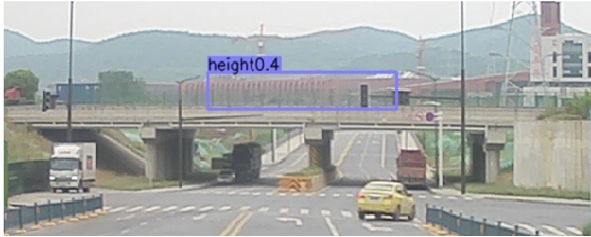}}
	\subcaptionbox{\centering After Extension\label{fig: After Extension}}
	{\includegraphics[width=0.49\linewidth]{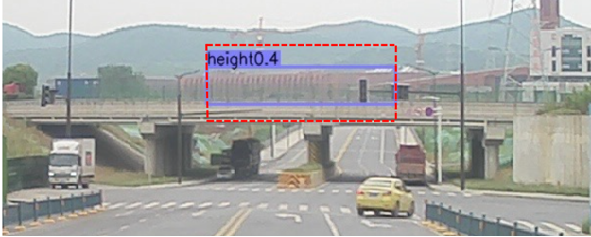}}
	\caption{Pixel Extension}\label{Pixel Extension}
\end{figure}

\par
Firstly, we conduct pixel extension. We extend the lower boundary of the predicted box, then execute frustum-based target extractor for the extended points, finally back-project them into a point cloud. For height limit estimation task, it is important to obtain the lower edge line of the device if the height in 3D space can be accurately calculated. Existing object detection methods, even if they can localize the objects, can not label all the object contours as in semantic segmentation. Object detection pay more attention to the whole detected object instead of guaranteeing prediction boxes completely contains the lower edge of the height limit device, we would rather contain more than miss one, as shown in Fig. \ref{fig: Before Extension}. Moreover, if the extension parts are noise, we still can filter by our spatial-based depth filter, in other word, pixel extension at least does not reduce accuracy. Thus, we should extend the lower boundary of the prediction bounding box of $M$ pixels as shown in Fig. \ref{fig: After Extension}, where $M$ is the hyper-parameter.

\par
Secondly, we conduct regulate effective interval. We can simply define the effective depth interval, i.e. $depth\in \left [ Z_{cen}-\sigma , Z_{cen}+\sigma \right ]$. The points inside are treated as effective points. The points outside can be treated as noise, and will be removed, we let interval radius $\sigma$ as the hyper-parameter of depth filter, and interval center point $Z_{cen}$ determined by Kernel Density Estimation\cite{kernel_density_estimation}.

\begin{figure}[!h]
  \centering
  \includegraphics[width=0.49\textwidth]{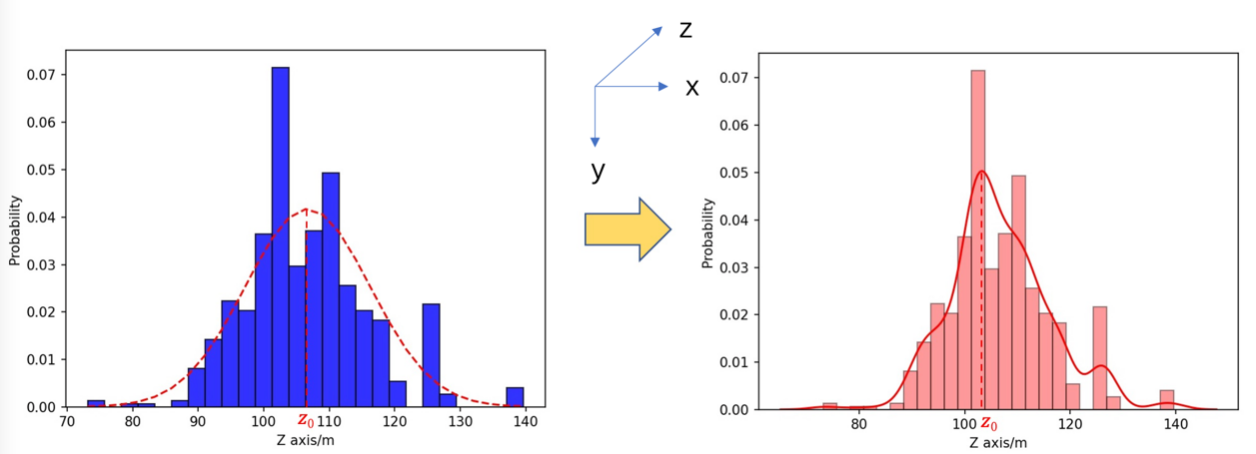}
  \caption{Kernel Density Estimation}
  \label{fig: Kernel Density Estimation}
\end{figure}

\par
Third, we conduct kernel density estimation \cite{kernel_density_estimation}. We first believe that the probability distribution of depth of the points in object detection boundary box should be similar to the normal distribution, that is, bell-shaped, low at the ends and high in the middle. Thus, we first treat its distribution as normal distribution with big standard deviation. Then, interval center point $Z_{cen}$ is the mean of all points. However, the effect does not work as hoped, as left image of Fig. \ref{fig: Kernel Density Estimation} show. The main reason is that the appearance of noise part often locates in either small value area or large value area, instead of appearing simultaneously.

\par
Thus, we set our sights on Kernel Density Estimation(KDE)\cite{kernel_density_estimation} to better fit our data. A natural idea is that if we want to know the value of the density function at $X = x$, we can pick a small interval around x, as in a histogram, and count the number of points inside that interval, divided by the total number. Likewise, KDE treat all the samples as neighbours, and the near neighbours make large influence, far neighbours make small influence, and kernel function $\varphi (x)$ is used as the metric to measure the distance and the influence of neighbours. Strictly speaking, KDE is a non-parametric estimation method, which are used to fit the observed sample points, as right image of Fig. \ref{fig: Kernel Density Estimation} shows. The depth value, corresponding to maximum probability density of KDE probability distribution curve, is the interval center point $Z_{cen}$. Due to the excellent mathematical properties of Gaussian function, we chosen it as the kernel function in our task. Bandwidth $h$ is the hyper-parameter, and regulates jitter degree in kernel function.

\begin{equation}
\displaystyle
\hat{f}(x)=\frac{1}{h N_{\text {total}}} \times \sum_{i=1}^{N_{\text {total}}} \phi\left(\frac{x-x_{i}}{h}\right)
\label{eq: Kernel Density Estimation}
\end{equation}

\subsubsection{Temporal-Based Height Filter}
\par
The vehicle is subject to road bumps during travel, so the data collected at each moment can not be exactly the same. Specifically, after the calculation by SHLE, each predicted height of the height limit device may not be the same for each frame of the same scene. Therefore, it is necessary to keep predicted height values within a relatively stable range of variation. Thus, we design a temporal-based height filter inspired by Kalman Filter\cite{kalman_filter}.

\par
Firstly, we average the lowest points in the point cloud. Height of the height limit device is in fact the minimum value in y-axis of the corresponding points. However, some avoidable errors may occur during objection detection, depth acquisition or 2D-3D project process. Thus, we calculate the average height of the lowest $N$ points to reduce errors to a degree.

\begin{figure}[!h]
	\centering
	\subcaptionbox{Kalman in image014 scene\label{fig: Kalman image014}}
	{\includegraphics[width=0.49\linewidth]{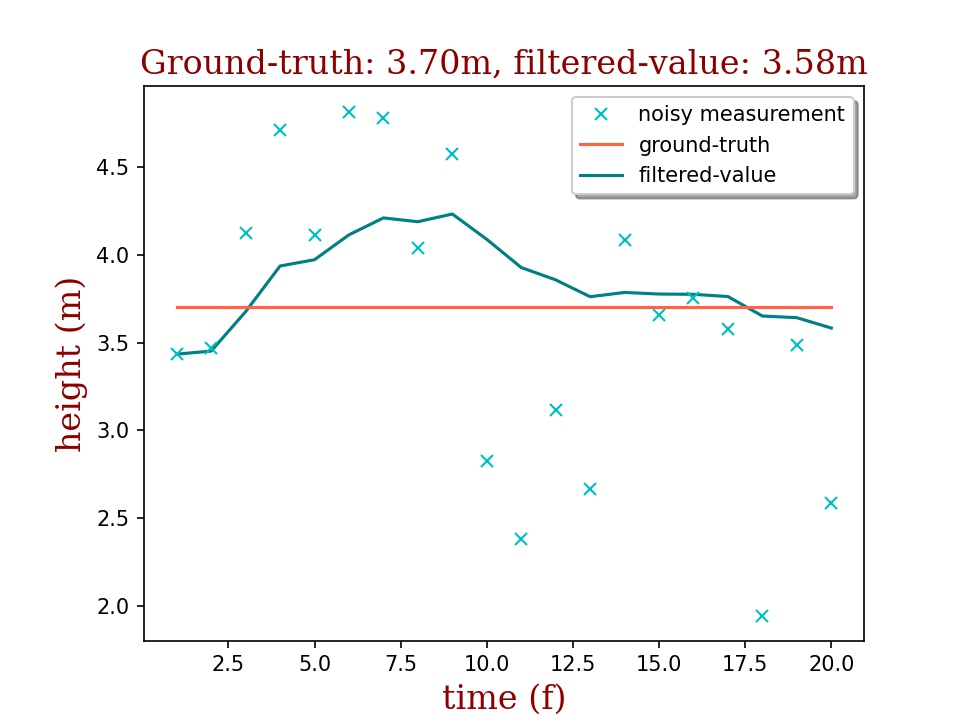}}
	\subcaptionbox{Kalman in image016 scene\label{fig: Kalman image016}}
	{\includegraphics[width=0.49\linewidth]{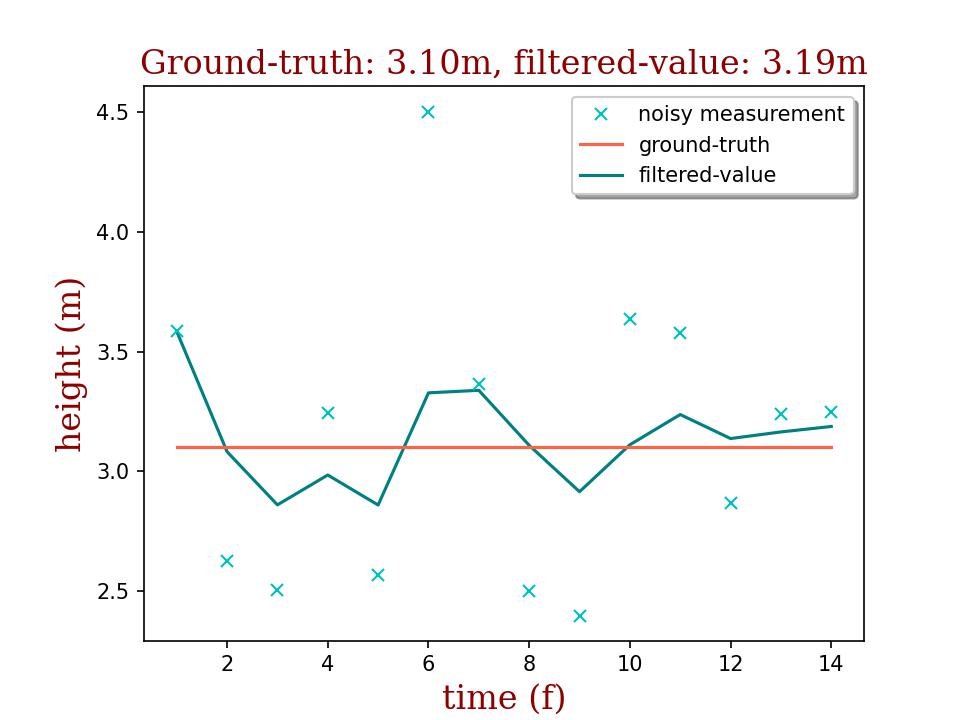}}
	\caption{Kalman Filter}\label{Kalman Filter}
\end{figure}

\par
Secondly, we tune the height in a video by Kalman filter \cite{kalman_filter}. Kalman filter\cite{kalman_filter} can make our time series smoother and attenuate the effect of noise, while also integrating previous time series estimations. Kalman filter\cite{kalman_filter} has been proved to be the optimal estimation in linear sequence problems. Kalman filter\cite{kalman_filter} adjust the height prediction series for a height limit device, as shown in Fig. \ref{Kalman Filter}. Variance of transfer noise is $Q$ and variance of measurement noise is $R$, and they are treated as hyper-parameters in our task.

\subsection{Evaluation Metrics} \label{evaluation metric}
\par
As SHLE is divided into two stages, we have different evaluation metrics for the output of the two stages, which we present in turn.

\subsubsection{Metrics for Stage 1}

\begin{figure}[!h]
	\subcaptionbox{\centering \label{fig: metric a}}
	{\includegraphics[width=0.49\linewidth]{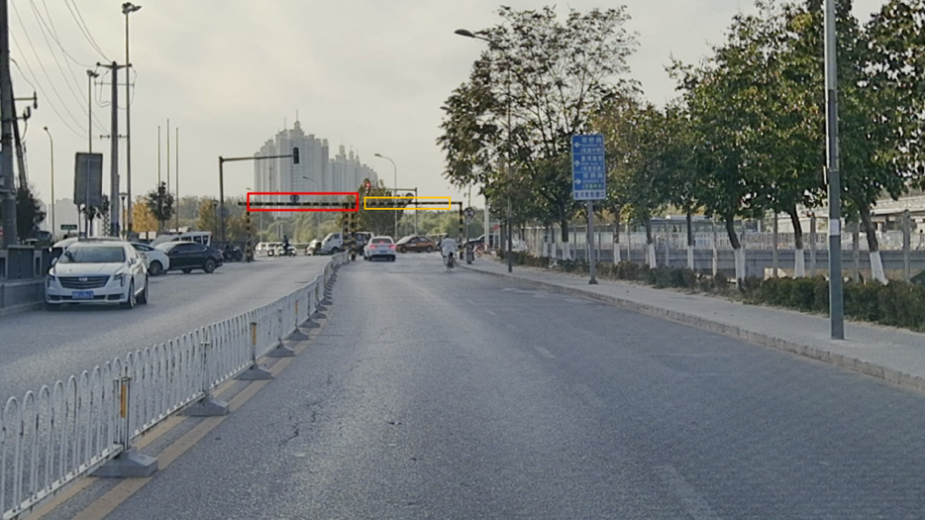}}
	\subcaptionbox{\centering \label{fig: metric b}}
	{\includegraphics[width=0.49\linewidth]{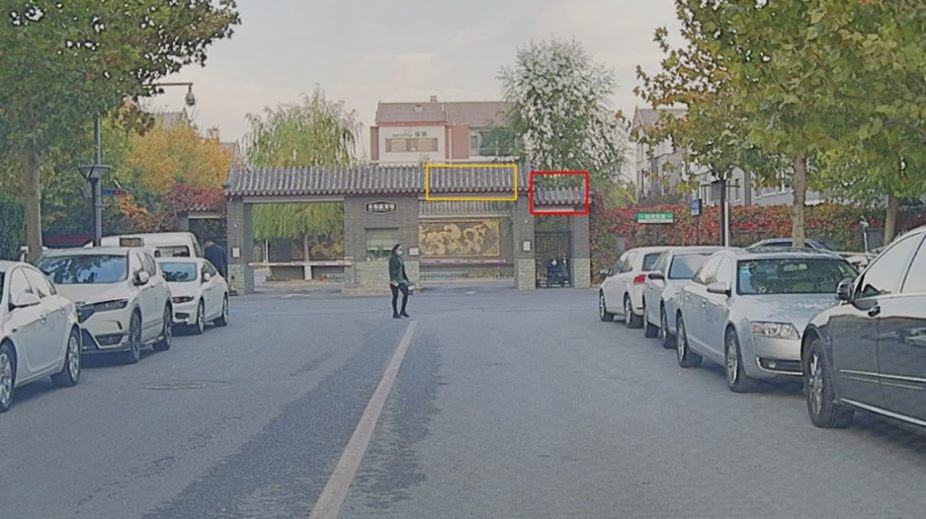}}
	\caption{Red means prediction detection box, yellow means ground truth\centering}\label{metric}
\end{figure}

\par
For conventional datasets such as MSCOCO or ImageNet, the commonly used evaluation metrics for object detection are AP and mAP. However, we find these metrics are not suitable for our height limit devices detection task. In some cases, though the metric demonstrate good values, it cannot reflect the true performance of the models. For example, as Fig. \ref{fig: metric a} shows, although the IoU of the prediction box and ground truth is 0, and the AP and mAP are also 0, they still point to the same height limit device, and the final height estimation results are correct. Another example is shown in Fig. \ref{fig: metric b}, where the bounding boxes are correct while the evaluation metrics are not high. Thus, conventional metrics for object detection are obviously unsuitable in our task.

\par
Therefore, as Eq. \ref{eq: OD} show, we decide to adopt $CPD$ (center point distance), $RCPDA$ (the ratio of center point distance to the area of ground truth box), and $RCPDH$ (the ratio of center point distance to the hypotenuse of ground truth box) as evaluation metrics. The center point coordinate of prediction and ground truth are $p_{pred}(x_{pred}, y_{pred})$ and $p_{gt}(x_{gt}, y_{gt})$, and the width and height of ground truth are $w_{gt}$ and $h_{gt}$. The proposed evaluation metrics can be calculated as:

\begin{equation}
\begin{array}{c}
CPD = \sqrt{(x_{pred}-x_{gt})^2+(y_{pred}-y_{gt})^2} \\ \\
RCPDA = \displaystyle{\frac {CPD} {(w_{gt} \times h_{gt})}}  \\ \\
RCPDH = \displaystyle{\frac {CPD} {\sqrt{w_{gt}^2+h_{gt}^2}}}  \\
\end{array}
\label{eq: OD}
\end{equation}

\subsubsection{Metrics for Stage 2}
\par
For height prediction task, as Eq. \ref{eq: HE}, we simply adopt $HE$(Height error: the predicted height value minus the true height value of the height limiting device), $HER$(Height error rate: the ratio of height error to ground truth value) as evaluation metrics. The prediction height is $PH$, the ground truth of height is $GT$. The metrics can be calculated as:

\begin{equation}
\begin{array}{c}
HE = PH - GT \\ \\
HER = \displaystyle{\frac{PE}{GT}}  \\
\end{array}
\label{eq: HE}
\end{equation}

\section{Experiment}
\subsection{Dataset and Metric}

\begin{figure}[!h]
  \centering
  \includegraphics[width=0.49\textwidth]{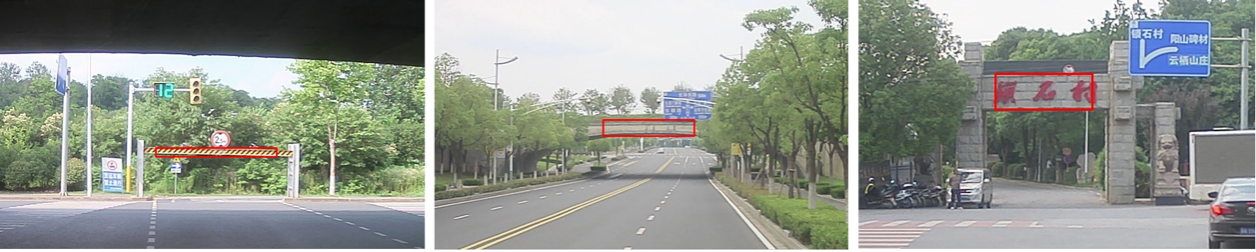}
  \caption{Data Annotation}
  \label{fig: Data Annotation}
\end{figure}

\par
Since we are the first to utilize vision based methods for height limit estimation task. Therefore, there is no public available dataset that we can use. To benchmark our task, we propose a novel large-scale dataset named "Disparity Height". For shooting setting, the baseline of our stereo camera is 120 mm, the camera mounting height is 1.45 m, the resolution of taken photos is $1280 \times 720$. "Disparity Height" is collected in natural outdoor driving conditions and has a total of 1587 images from 13 scenes, which covers large different types of height limit devices. Simultaneously, corresponding height of the limiters, camera parameters and stereo disparity map are also recorded. Then each image is labelled with bounding box of corresponding height limit device by $Labelme$ as Fig. \ref{fig: Data Annotation}. In this dataset, we not only consider different scenes, but also consider the effects of different lighting, weather, etc. Moreover, to verify the generalizability of SHLE designed in this paper, we use the data in the first 10 scene as the training set, and the data in the last 3 scene as the test set. Therefore, "Disparity Height" is a very challenging dataset. It is very meaningful to evaluate the performance of our method and the baselines.

\par
As mentioned in  section \ref{evaluation metric}, to better accommodate height estimation task, we take $CPD$, $RCPDA$, and $RCPDH$ as the main evaluation metrics for stage 1, that is, object detection and tracking step. And then we take $HE$ and $HER$ as the evaluation metrics for stage 2, that is, height estimation step.

\subsection{Main Results}

\subsubsection{Baseline Comparison}
\par
Since nobody currently uses deep learning methods in height estimation tasks through front view, we can only transfer models that have achieved excellent results in other similar tasks as baseline for experimental effect pairs, as shown in TABLE \ref{table: baseline}.

\begin{table}[h]
    \begin{minipage}{0.98\linewidth}
    \centering	
    \caption{Height Estimation Baseline Comparison.}
    \label{table: baseline}
    \resizebox{0.6\linewidth}{!}{
        \begin{tabular}{lccc}
            \hline \hline
            Model & HER & HE \\
            \hline
            MobileNetV2 \cite{mobilenetv2} & 10.77 & 0.32 \\
            UNet \cite{unet}	    &  8.79 & 0.26 \\
            DORN \cite{dorn}	    & 14.03 & 0.42 \\
            ViT  \cite{vision_tranformer}     	& 11.18	& 0.34 \\
            CoAN \cite{canet}	    & 14.23	& 0.43 \\
            PointNet \cite{pointnet}	& 18.86	& 0.65 \\
            \hline
            \textbf{Our Best}    & \textbf{2.67}	& \textbf{0.08} \\
            \hline \hline
        \end{tabular}
    }
    \end{minipage}
\end{table}

\par
Image classification and height estimation task have some similarities that they both take the whole image as input, then extract a overall feature, and finally output a global result. Thus, we can apply corresponding model by modifying the last classification layer to a linear regression layer in this paper. MobileNetV2\cite{mobilenetv2} is a lightweight neural network that has achieved excellent results in image classification tasks. MobileNetV2\cite{mobilenetv2} is 8.10\% in HER and 0.24m in HE worse than our best. We believe that MobileNetV2\cite{mobilenetv2} pays more attention to global information instead of understanding local information near key parts, thus easily affected by the noise, that is, non-height-limit device part. Vision Transformer(ViT)\cite{vision_tranformer} combines CV and NLP domain knowledge without relying on CNN and recently has achieved good results on image classification tasks. ViT is 8.51\% in HER and 0.26m worse than our best. The resolution of our input image is $1280 \times 720$, which is too computationally intensive for ViT\cite{vision_tranformer}. Thus, we have to resize the image in lower size $480 \times 270$, but resize process avoidably loss too much effective information, and still too large to handle.

\par 
Monocular depth estimation and height estimation task are a both regression problem for a image, though the former is at pixel level and the latter is at image level. Thus, we can apply corresponding model by weighted averaging the output of all pixels with a attention layer. Deep Ordinal Regression Network(DORN)\cite{dorn} is a lightweight network for monocular depth estimation. DORN\cite{dorn} is 11.36\% in HER and 0.34\% in HE worse than our best. The ASPP of DORN\cite{dorn} does not compute all input pixels, i.e., the convolution kernel is not continuous. The use of ASPP may be effective for large objects, but may be counterproductive for small objects, such as our height limit device.

\par
UNet is originally a lightweight semantic segmentation network with a wide range of applications in biomedical image analysis, UNet\cite{unet} family are extended to be used in regression task in other area recently. In our task, we add a attention layer, and weighted average the output of original UNet. UNet\cite{unet} is 6.12\% in HER and 0.18 in HE worse than our best. Although UNet is just a simple encoder-decoder network with residual connect, it acquire the best result in our baseline. Because of the attention mechanism, A-UNet\cite{unet} can simply distinguish the key part and the unrelated part, but is still a little behind.

\par
Since we hold both image and depth data (disparity), the models that act on 3D data(e.g. voxels, point clouds, RGB-D images) can take more information as input, might work better in our height estimation task. Co-Attention Network(CANet)\cite{canet} achieves RGB and depth effective fusion with a long range fusion instead of local fusion. Since CANet\cite{canet} is originally used for segmentation task, we modify it as what we do for MobileNetV2\cite{mobilenetv2}. CANet\cite{canet} is 11.56\% in HER and 0.35m in HE worse than our best. The fusion strategy of CANet\cite{canet} is not suitable for our task, it seemingly deep fusion, but does not represent effect. PointNet\cite{pointnet} is used for processing point cloud data, and its status is comparable to that of CNN networks in 2D image processing. Since PointNet\cite{pointnet} is originally used for segmentation and classification task, we modify it as above MobileNetV2\cite{mobilenetv2}. PointNet\cite{pointnet} is 16.19\% in HER and 0.57m in HE worse than our best. The main reason of the bad result is that our point clouds just projected from a RGB-D image, missing much 3D information. Thus, regular point cloud process methods are not applicable. 

\begin{figure}[!h]
	\subcaptionbox{\centering Scene Imagess017\label{fig: demo0}}
	{\includegraphics[width=0.98\linewidth]{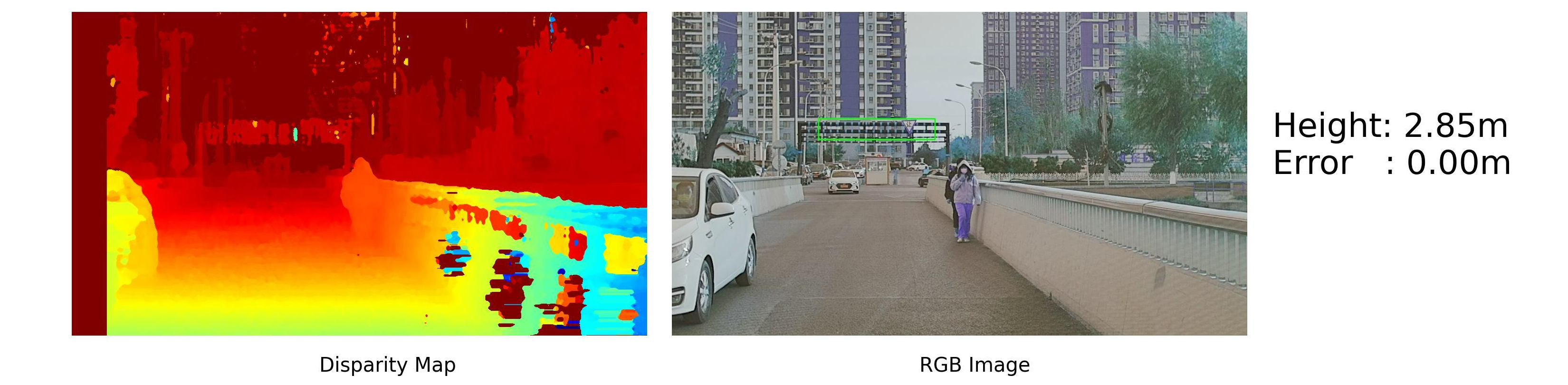}}
	\subcaptionbox{\centering Scene Imagess018\label{fig: demo1}}
	{\includegraphics[width=0.98\linewidth]{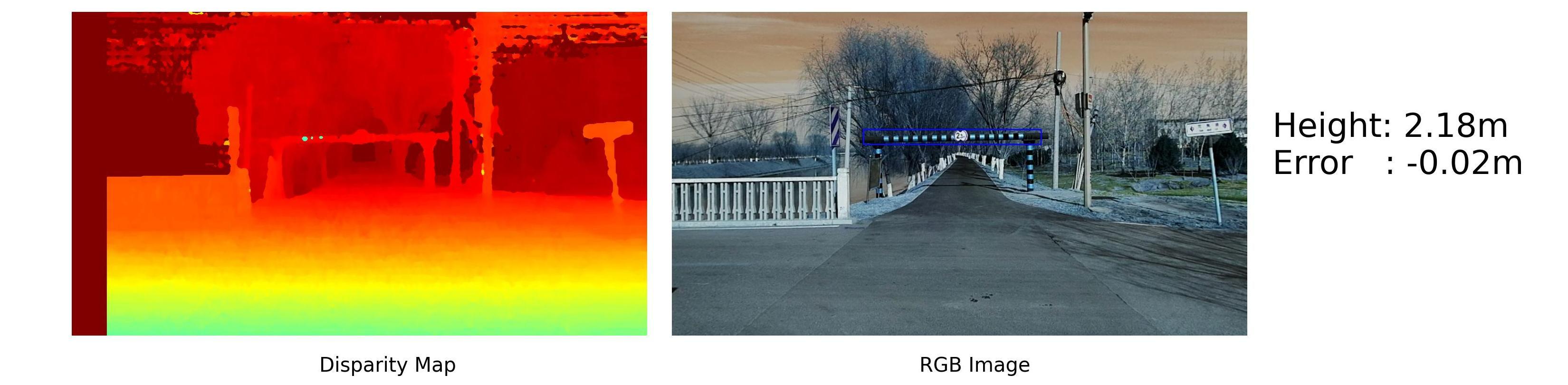}}
	\subcaptionbox{\centering Scene Imagess019\label{fig: demo2}}
	{\includegraphics[width=0.98\linewidth]{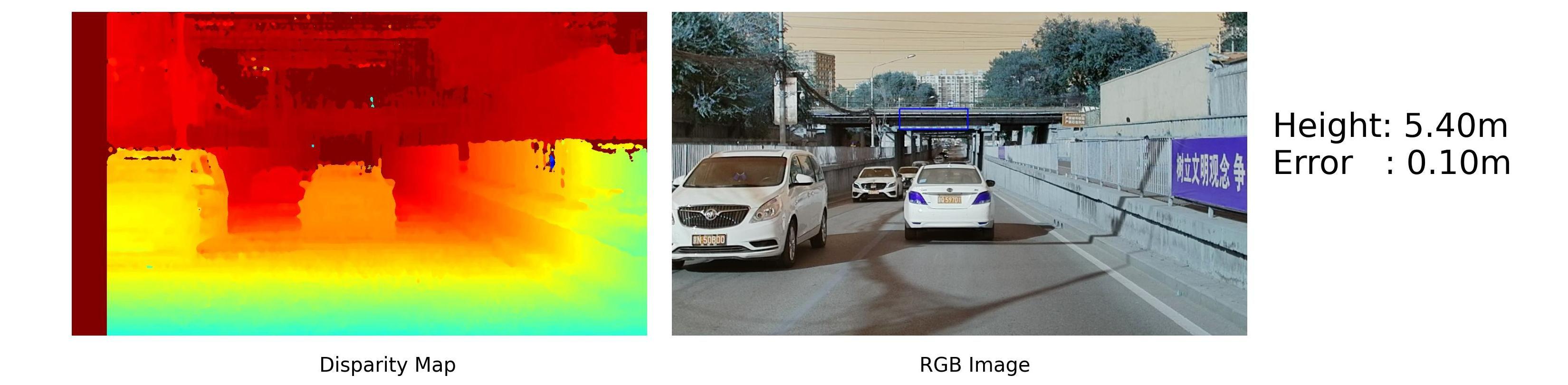}}
	\caption{visualisation effect of SHLE\centering}\label{fig: demo}
\end{figure}

\par
Our trained model is executed on demo data rather than training or validation data, Fig. \ref{fig: demo} shows the visualisation effect of SHLE. As you can see, our method has little error and is able to be applied in practice. See supplementary material for the demo video.

\subsubsection{Hyper-parameters}
\par
Hyper-parameters will be introduced in order.

\begin{table}[h]
    \begin{minipage}{0.98\linewidth}
    \centering	
    \caption{Object Detection Method Comparison.}
    \label{table: Object Detection}
    \resizebox{\linewidth}{!}{
        \begin{tabular}{lccc}
            \hline \hline
            Model & CPD	& RCPDA	& RCPDH \\
            \hline
            Yolov5 \cite{yolov5}   & 131.21	& 0.043	& 0.65 \\
            CenterNet \cite{centernet} & 122.05	&0.45	& 0.82 \\
            RetinaNet \cite{retinanet} & 165.27 	&0.037	& 0.86 \\
            FCOS    \cite{fcos}	  &188.19	&0.029	& 0.77 \\
            Faster RCNN \cite{faster_rcnn}	&113.49	&0.034	& 0.67 \\
            \hline
            \textbf{Faster RCNN+Filter Rule} & \textbf{95.74} & \textbf{0.025} & \textbf{0.55} \\
            \hline \hline
        \end{tabular}
    }
    \end{minipage}
\end{table}

\par
In stage 1, the performance of height limit devices detector directly affects follow-up process, method of object detection can be treated as the backbone. Faster RCNN\cite{faster_rcnn} reaches the best results as TABLE \ref{table: Object Detection}. The RoI pooling of Faster RCNN\cite{faster_rcnn} make it adapted to handle multi-scale object, such as the height limit device in our task.

\begin{figure*}[!h]
    \centering
    \subcaptionbox{\centering Pixel Extension\label{fig: Pixel Extension}}
    {\includegraphics[width=0.24\linewidth]{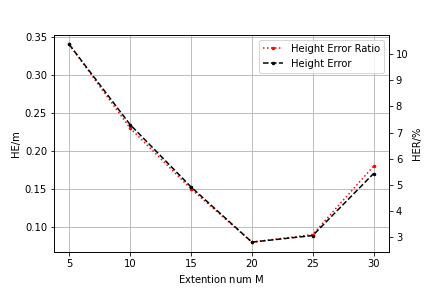}}
    \subcaptionbox{\centering Depth Filter\label{fig: Depth Filter}}
    {\includegraphics[width=0.24\linewidth]{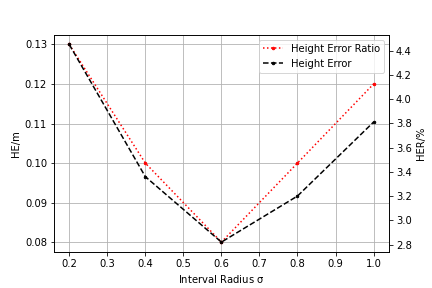}}
    \subcaptionbox{\centering Average Lowest Points\label{fig: Average Lowest Points}}
    {\includegraphics[width=0.24\linewidth]{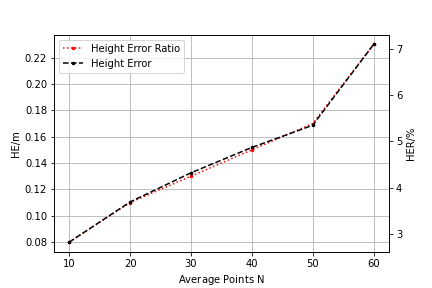}}
    \subcaptionbox{\centering Kernel Density Estimation\label{fig: Kernel Density Estimation1}}
    {\includegraphics[width=0.24\linewidth]{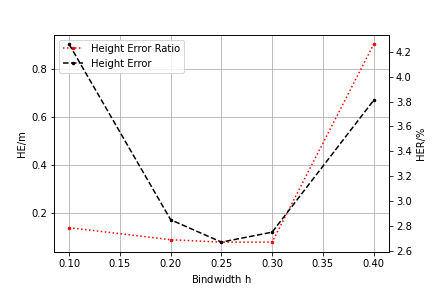}}
    \subcaptionbox{\centering Kalman Filter\label{fig: Kalman Filter Q}}
    {\includegraphics[width=0.24\linewidth]{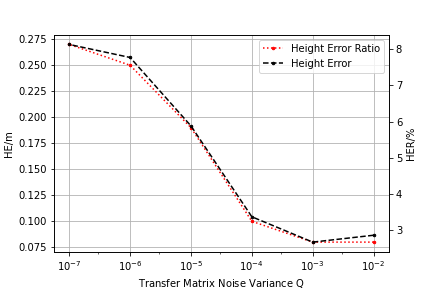}}
    \subcaptionbox{\centering Kalman Filter\label{fig: Kalman Filter R}}
    {\includegraphics[width=0.24\linewidth]{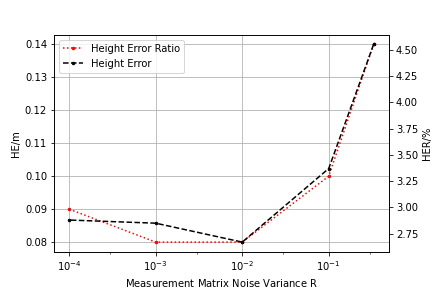}}
    \subcaptionbox{\centering Object Tracking\label{fig: Object Tracking}}
    {\includegraphics[width=0.24\linewidth]{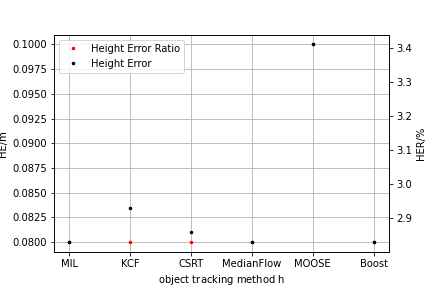}}
    \caption{Hyper-parameter. Fig. \ref{fig: Hyper-parameter} shows the specific hyper-parameter process, taking $HER$(red) and $HE$(black) as vertical axis.}
    \label{fig: Hyper-parameter}
\end{figure*}

\par
In SHLE, hyper-parameters of each step all have different impact. MIL reaches best results in chosen method of object tracking, as shown in Fig. \ref{fig: Object Tracking}, but it does not show significant superiority over other methods. $M=20$ reaches best results of the number of Pixel Extension, as Fig. \ref{fig: Pixel Extension}. $\sigma=0.6$ reaches best results of the interval radius of Depth Filter, as shown in Fig. \ref{fig: Depth Filter}. $h=2.5$ reaches best results of the bandwidth of Kernel Density Estimation\cite{kernel_density_estimation}, as shown in Fig. \ref{fig: Kernel Density Estimation}.  $N=10$ reaches best results of the number of Average lowest points, as shown in  Fig. \ref{fig: Average Lowest Points}. $Q=1e-3$ and $R=1e-2$ reach best results of the variance of transfer matrix and measurement matrix of Kalman Filter\cite{kalman_filter}, as shown in Fig. \ref{fig: Kalman Filter Q} and Fig. \ref{fig: Kalman Filter R}.

\subsection{Ablation Study}
\begin{table}[h]
    \begin{minipage}{0.98\linewidth}
    \centering
    \caption{Ablation Study.}
    \label{table: Ablation Study}
    \resizebox{0.8\linewidth}{!}{
    \begin{tabular}{lccc}
            \hline \hline
            Model & HER & HE \\
            \hline
            -Pixel Extension & 14.47 & 0.47 \\
            -Depth Filter & 34.01 & 1.1 \\
            -Kernel Density Estimation	& 16.72	& 0.52 \\
            -Average Lowest Points & 3.18 & 0.1 \\
            -Kalman Filter	& 2.82 & 0.08 \\
            \hline
            \textbf{Our Best model} & \textbf{2.67}	& \textbf{0.08} \\
            \hline \hline
        \end{tabular}
    }
    \end{minipage}
\end{table}

\subsubsection{Effect of Novel Filter Rule}
\par
Although objection detection has already taken NMS(Non-Maximum Suppression) for filtering the outputs, few detection bounding box still remain, but we only need one in our task. Therefore, we make a few assumptions based on the characteristics of height limit devices in outdoor driving scene, and design novel filter rules based on this assumptions instead of just simply choosing the one that has the highest confidence. TABLE \ref{table: Object Detection} shows the improvement of applying novel filter rule, achieving 17.75 better in $CPD$, 0.009 better in $RCPDA$, and 0.12 better in $RCPDH$. The experimental results shows the validity of our novel filter rules.

\subsubsection{Spatial-Based Depth Filter}
\par
Since spatial-based depth filter is divided into three part: pixel extension, depth filter, and kernel density estimation\cite{kernel_density_estimation}. We will explore their impact separately.

\par
Effect of Pixel Extension. Although our height limit estimation task just cares its lower edge, but object detection focus on integrity of object, which perhaps neglect the lower edge information. Even the pixel extension may introduce some noise, later depth filter also can remove it, we would rather contain more than miss one. TABLE \ref{table: Ablation Study} shows the improvement of applying Pixel Extension, achieving 11.80\% better in $HER$ and 0.39m better in $HE$. The experimental results shows the validity of pixel extension method.

\par
Effect of Depth Filter. Although we just focus on the height limit device in a image, the noise is unavoidably contained in our bounding box, unlike semantic segmentation task. The presence of the noise, such as tree, wire poles, billboards, part seriously interferes with the calculation of final minimum height. Meanwhile, the depth of noise always has a large gap with the effective part, we can design a depth filter to remove it. TABLE \ref{table: Ablation Study} shows the extreme improvement of applying Depth Filter, achieving 31.34\% better in $HER$ and 1.02m better in $HE$. The experimental results shows the validity of depth filter method, depth filter indeed remove the noise, and can be extend to other tasks for height limit device.

\par
Effect of Kernel Density Estimation\cite{kernel_density_estimation}. Depth Filter extremely improve the experimental result, how to determine the interval center point $Z_{cen}$ of depth filter interval matters a lot. We first assume the depth in bounding box belongs to the normal distribution, that is, the mean of depth is $Z_{cen}$, but TABLE \ref{table: Ablation Study} shows the bad result. The main reason is that the depth of noise, such as sky, is too large compared to effective part, getting results by mean value will make $Z_{cen}$ heavily influenced. Thus, we use kernel density estimation\cite{kernel_density_estimation} to address this problem, TABLE \ref{table: Ablation Study} shows the great improvement of applying Kernel Density Estimation, achieving 14.05\% better in $HER$ and 0.44m better in $HE$. The experimental results shows the validity of kernel density estimation\cite{kernel_density_estimation}. 

\subsubsection{Temporal-Based Height Filter}
\par
Since temporal-based Height filter is divided into two part: Average Lowest Points and Kalman Filter\cite{kalman_filter}. We will explore their impact separately.
\par
Effect of Average Lowest Points. Due to the error and noise in disparity map and object detection, we infer that only choosing the lowest point as the output is not the reasonable strategy. TABLE \ref{table: Ablation Study} shows the little improvement of applying average lowest points, achieving 0.51\% better in $HER$ and 0.02m better in $HE$. The experimental results shows the validity of average lowest points. 

\par
Effect of Kalman Filter\cite{kalman_filter}. The vehicle is affected by road bumps during driving, and the collected data can not be perfectly consistent at each moment. Therefore, we need to use Kalman filter to keep the current predicted height within a relatively stable variation range. Since the accuracy has reached a high level, TABLE \ref{table: Ablation Study} just shows the little improvement of applying Kalman Filter\cite{kalman_filter}, achieving 0.15\% better in $HER$. The experimental results shows the validity of Kalman filter\cite{kalman_filter}.

\subsection{Future Direction}
\par
The height estimation pipeline SHLE we design in this paper for height limit device is only the tip of the iceberg in the field of autonomous driving, and there is much more knowledge to be explored.

\par
The road condition is complex and variable, and more scenes can be collected later. The scene should not only include height limit device with traffic signage commonly found on urban roads, but also need encompass non-urban, traffic signage free height limit device.

\par
In terms of assisted driving, the installation of LIDAR will ensure that the height estimation ability does not fail even in poor visual conditions. Thus, in the future, the integration of stereo cameras with LIDAR can be tried to explore whether the height prediction capability of height limit devices can be further improved, and to enrich the application scenarios of our model.

\par
Besides, currently, we assume there is only one height limit devices in a scene, but in some cases, there may be several different height limit devices in a scene. Therefore, we have to research this direction in the future.

\section{Conclusion}
\par
In this paper, we propose a two-stage height estimation pipeline SHLE that combines deep learning and traditional methods, and SHLE is the first time that deep learning is used for height limit estimation task. To benchmark the height limit estimation task, we build a large-scale dataset called "Disparity Height". Based on the characteristics of the height limit device and the nature of the height estimation task, we propose a category agnostic object detector and a category agnostic tracker in stage 1, and we create a frustum based extractor, spatial-based depth filter and a temporal-based height filter in stage 2. Our method currently achieves the best in this area, far exceeding other end-to-end deep learning methods in terms of the evaluation metrics of height estimation.


\ifCLASSOPTIONcaptionsoff
  \newpage
\fi



\printbibliography

\end{document}